\documentclass[10pt, conference, compsocconf]{IEEEtran}

\usepackage{times}
\usepackage{epsfig}
\usepackage{graphicx}
\usepackage{amsmath}
\usepackage{amssymb}
\usepackage{xcolor,colortbl}

\DeclareMathAlphabet{\mbf}{OT1}{ptm}{b}{n}
\newcommand{\mbs}[1]{{\boldsymbol{#1}}}
\newcommand{\norm}[1]{\left\Vert#1\right\Vert} % Norm
\newcommand{\abs}[1]{\left\vert#1\right\vert}

\newcommand{\mrm}[1]{\mathrm{#1}}
\newcommand{\trans}{{\ensuremath{\mathsf{T}}}}

\begin{document}

%%%%%%%%% TITLE
% \title{Class Instance Balanced Learning for Long-Tailed Classification}

% \author{Marc-Antoine Lavoie\\
% University of Toronto Institute for Aerospace Studies \\
% Toronto, Canada \\
% {\tt\small marc-antoine.lavoie@robotics.utias.utoronto.ca}
% \and
% Steven L. Waslander \\
% University of Toronto Institute for Aerospace Studies \\
% Toronto, Canada \\
% {\tt\small steven.waslander@robotics.utias.utoronto.ca }
% }

\def\IEEEbibitemsep{0pt plus .5pt}

\title{Class Instance Balanced Learning for Long-Tailed Classification}
\author{
Marc-Antoine Lavoie and Steven L. Waslander\\
Institute for Aerospace Studies \\ 
University of Toronto \\
Toronto, Canada \\
{\tt\small \{ marc-antoine.lavoie, steven.waslander \}@robotics.utias.utoronto.ca }
}

\maketitle

%%%%%%%%% ABSTRACT
\begin{abstract}
The long-tailed image classification task remains important in the development of deep neural networks as it explicitly deals with large imbalances in the class frequencies of the training data. While uncommon in engineered datasets, this imbalance is almost always present in real-world data. Previous approaches have shown that combining cross-entropy and contrastive learning can improve performance on the long-tailed task, but they do not explore the tradeoff between head and tail classes. We propose a novel class instance balanced loss (CIBL), which reweights the relative contributions of a cross-entropy and a contrastive loss as a function of the frequency of class instances in the training batch. This balancing favours the contrastive loss for more common classes, leading to a learned classifier with a more balanced performance across all class frequencies. Furthermore, increasing the relative weight on the contrastive head shifts performance from common (head) to rare (tail) classes, allowing the user to skew the performance towards these classes if desired. We also show that changing the linear classifier head with a cosine classifier yields a network that can be trained to similar performance in substantially fewer epochs. We obtain competitive results on both CIFAR-100-LT and ImageNet-LT.
\end{abstract}

%%%%%%%%% BODY TEXT
\section{Introduction}

The rapid development of convolutional neural network (CNN) architectures has allowed great progress in many computer vision tasks, such as classification, object detection and segmentation. However, moving from the hand-engineered benchmark datasets used in most image classification tasks to unfiltered datasets collected from real-world experiments presents a problem. Real datasets tend to be significantly more imbalanced, with some classes being much more common than others. Consider for instance nuImages \cite{caesar2020nuscenes}, an autonomous driving dataset obtained from real-world driving scenes, in which there are 36314 truck instances but only 42 ambulances. Standard training approaches will generate classifiers biased toward the most common classes when applied to this imbalanced dataset.

In long-tailed classification, we observe two main problems on the rarest classes: underfitting and overfitting. When underfit, the classifier does not learn to classify rare class examples during training and performs poorly on both the training and testing datasets. On the other hand, when overfit, the network performance is good on training examples but does not generalize to the unseen test data.

Common strategies to improve performance on the rare classes involve correcting for the data imbalance in the training signal, either by adjusting the sampling by undersampling frequent classes \cite{he2009learning} or oversampling rare classes \cite{huang2016learning, mahajan2018exploring}, or by balancing the gradient of the loss by weighting the individual losses according to class frequencies \cite{khan2017cost,cui2019class}. These methods generally trade performance on the most frequent classes to improve results on the rarer ones \cite{buda2018systematic}.

A refinement of the oversampling approaches involves the generation of new samples, particularly from the rare classes, to correct the initial imbalance. These new samples can be generated in the image space using more robust augmentations \cite{yun2019cutmix} or generative models \cite{li2020adversarial}, or in feature space by transforming existing features with the addition of noise \cite{li2021metasaug}. Similarly, some developments on the loss gradient balancing approach are the margin \cite{cao2019cifarlLT} and logit adjustment methods \cite{ren2020balanced}. Instead of simply increasing the magnitude of the gradient for rare classes, the training task is made harder and more robust for these classes.

Recently, there has been an interest in using contrastive losses \cite{chen2020simple, chen2020improved} instead of the standard cross-entropy formulation. These losses have been shown to be effective in long-tailed classification, learning more balanced feature representations \cite{kang2021exploring}. Some approaches \cite{wang2021contrastive, cui2021parametric, zhu2022balanced} now combine contrastive and cross-entropy terms and obtain state-of-the-art performance. 
% Another development is the use of attention and transformer architectures. Vision transformers are now state-of-the-art for the standard classification tasks, but they have lagged on the long-tailed problem. However, some implementations have successfully used attention mechanisms when implemented on the final layers of a standard convolutional network.

Following the strong results shown by PaCo \cite{cui2021parametric}, we propose a similar structure combining cross-entropy and supervised contrastive losses. In contrast to that approach, we explicitly decouple the two losses and thus consider a simple weighted sum of both. The weights bias the loss function towards the contrastive loss for classes with many examples in the batch and momentum queue. We show that this reweighting, and not simply the sum of the two losses, improves performance and gives the user a single parameter to tune the network's performance towards the rarest classes. Next, we show that replacing the final linear classifier with a cosine classifier yields a network with competitive performance that requires significantly fewer epochs to train. Our main contributions are:
\begin{itemize}
    \item We show that when combining cross-entropy and contrastive losses, increasing the relative importance of the contrastive loss on the head classes effectively balances the performance across all classes, and this can be done by tuning a single value.
    \item Next, we show that simply modifying the linear classifier to a cosine classifier gives us a similar performance with much fewer training epochs.
    \item The proposed approach achieves competitive results on two popular long-tailed benchmarks.
\end{itemize}

% combination of two methods to improve performance on the long-tailed data. First, we show that a combination of cross-entropy and supervised contrastive losses, specifically one that forces frequent classes to be more sensitive to the contrastive loss, can improve performance on tail classes with little impact on head classes. Next, we show that the simple addition of a self-attention transformer module applied on the last layer of the CNN can significantly improve performance on rare classes.

%-------------------------------------------------------------------------

\section{Related Works}
\subsection{Long-Tailed Classification}
\subsubsection{Resampling and Reweighting} Classical approaches in long-tailed classification rely on directly compensating for the skew in training data. This can be solved by undersampling common classes \cite{he2009learning} or oversampling rare classes \cite{huang2016learning, mahajan2018exploring}, but the first can lead to overfitting while the second can lead to poor representation learning. Similar issues are observed with the direct loss reweighting approaches \cite{lin2017focal, khan2017cost, cui2019class}.

\subsubsection{Decoupled Learning} To reduce these effects, LDAM \cite{cao2019cifarlLT} proposes a two-stage training process, decoupling feature and classifier learning and ensuring that good features are learned before the balanced classifier is trained in a second step, using a more balanced sampling or weighting. In addition to simply training using a more balanced distribution, some methods \cite{kang2019decoupling, zhang2021distribution} consider post hoc adjustments that attempt to find an optimal reweighting of the logits to improve the final performance. DLSA \cite{xu2022constructing} trains a Gaussian mixture flow model as a second step to detect and send the tail instances to a separate classifying branch. 
% BBN \cite{zhou2020bbn} is essentially a continuous implementation of the same strategy, gradually shifting the loss towards a branch trained with oversampled tail instances.

\subsubsection{Margins and Logit Adjustment} Instead of scaling the loss according to class frequencies, some approaches instead attempt to make the classification task harder for instances of rare classes. This can be done by subtracting an offset from the true logit during training \cite{cao2019cifarlLT,deng2019arcface}, forcing the classifier to separate classes by an additional margin, which scales inversely with the number of instances in a class. Alternatively, logit adjustment methods \cite{ren2020balanced,menon2020long} opt instead to rescale all logits during training such that the most frequent ones are larger, again making the task more difficult for tail instances. ALA \cite{zhao2022adaptive} uses the instance difficulty in addition to class frequency to scale the added margins. 

\subsubsection{Additional Augmentations and Sample Generation} 
Some approaches use stronger augmentations or attempt to generate entirely new images to deal with the long tail. M2M \cite{kim2020m2m} transforms frequent class images by pushing them towards rare classes in image space, generating new rare images. CMO \cite{park2022majority} generates new tail samples by mixing tail and head images using Cutmix \cite{yun2019cutmix}. GAMO \cite{mullick2019generative} and AFHN \cite{li2020adversarial} generate new images using a GAN. MetaSAug \cite{li2021metasaug} perturbs the feature representation of rare class examples according to a learned covariance. A recent approach, OPeN \cite{zada2022pure}, adds pure noise images to the original instances from rare classes.

% Prototype-based approaches learn class-specific prototypes to help train more consistent classifiers or to transfer information from frequent to rare classes \cite{liu2019largeIN-LT-PL-LT, parisot2022long}.

% Self-attention, particularly with the use of vision transformers **cite*** has shown great promise in image classification and is now state-of-the-art. Performance has lagged on long-tailed classification, especially on the smaller datasets, but attention modules have been used successfully. OLTR uses a two-step process. In few-shot learning, attention over 
% OLTR, BAtchformer, attention paper

% \textbf{Sample Generation.} 
% ***To correct for this lack of variance, some methods consider injecting noise on the final logits \cite{liu2020deep}, while others try to generate more varied samples. This can be done by adding noise to the images \cite{li2020adversarial}, by combining multiple images together \cite{yun2019cutmix,park2022majority} or by interpolating in the feature space between two samples \cite{ko2020embedding}. A more recent approach simply adds pure noise images to the rare classes \cite{zada2022pure}. 

\subsubsection{Ensemble Methods} Long-tailed learning can also be improved using an ensemble of models instead of a single one. In RIDE \cite{wang2020long}, multiple models are trained with a diversity maximization loss, ensuring the models generate different outputs, and are deployed sequentially. LFME \cite{xiang2020learning} and NCL \cite{li2022nested} train multiple experts and use knowledge distillation to improve performance on a single model. ResLT \cite{cui2022reslt} uses a shared backbone but adds additional parameters to the heads trained on tail classes.

\subsection{Contrastive Learning} \label{sec:rel_cl}
\subsubsection{Pure Contrastive} Contrastive learning is an alternative to cross-entropy training originally developed for self-supervised training without labels \cite{he2020momentum, chen2020simple, chen2020improved}. These methods learn features by clustering different augmented views of the same image while pushing away all other images. Some works have shown that self-supervised contrastive pre-training can improve performance \cite{yang2020rethinking, li2021self} on the long-tailed task. Supervised Contrastive Learning (SCL) \cite{khosla2020supervised} is an extension of self-supervised contrastive learning when labels are available. It has been used successfully for classification on the full ImageNet dataset. However, this approach still suffers from issues with the long-tailed task and produces skewed results. To remedy this, KCL \cite{kang2021exploring} proposes to use a fixed number of positive samples, reducing the imbalance in training. TSC \cite{li2022targeted} adds fixed targets in the contrastive loss to act as class centers, ensuring a uniform class distribution in the feature space.
\subsubsection{Mixed Cross-Entropy and Contrastive Losses} Other approaches consider a mix of supervised contrastive and cross-entropy losses. Hybrid-SC \cite{wang2021contrastive} applies the loss in two independent branches and gradually decays the contrastive component of the loss in favour of the cross-entropy, allowing the network to first learn good features and then optimize the classifier for classification performance. PaCo \cite{cui2021parametric} adds the cross-entropy directly to the contrastive loss as additional terms, while BCL \cite{zhu2022balanced} first rescales the contribution of each class in the denominator of the contrastive loss and then adds learnable class centers derived from the cross-entropy classifier vectors.

%-------------------------------------------------------------------------

% \newpage

%-------------------------------------------------------------------------
\begin{figure*}[h]
    \centering
	\includegraphics[width=0.92\textwidth]{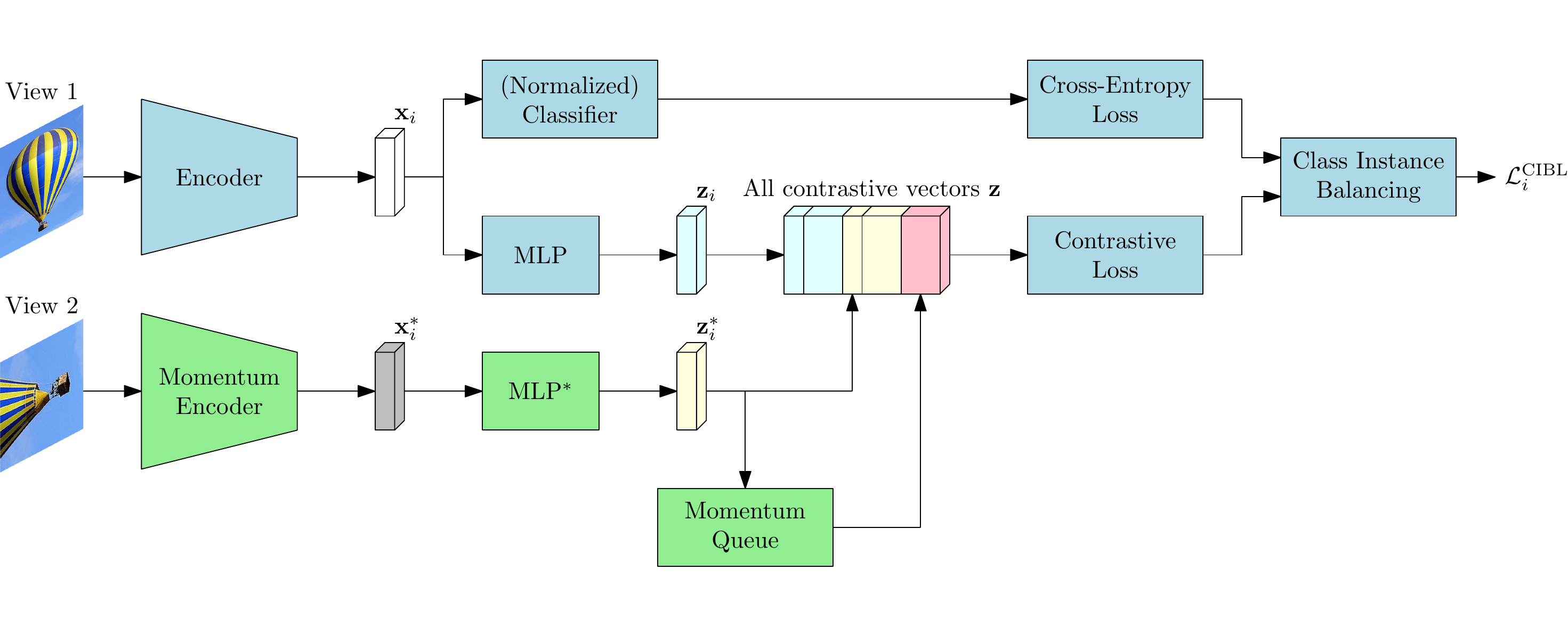}
	\caption{Class Instance Balanced Loss. Items marked with an $^{*}$ or in green are from the momentum branch.}
	\label{fig:schematic_CIBL}
\end{figure*}

\section{Method}
\subsection{Logit Adjustment} \label{sec:log_adj}
Balanced Softmax \cite{ren2020balanced} is a variation of the standard cross-entropy (CE) loss that adjusts the logits during training to account for the imbalance in the training set. The loss for instance $i$ is
\begin{align}
    \mathcal{L}_i^{\mrm{CE}} &= -\log \frac{n_{c_i} \exp \left(\mbf{x}_{i}^{\trans}\mbs{\theta}_{c_i} \right)}{\sum\limits_{c \in C} n_{c} \exp \left(\mbf{x}_{i}^{\trans}\mbs{\theta}_{c} \right)} \nonumber \\
    &= -\log \frac{\eta_{i,c_i}}{\sum\limits_{c \in C} \eta_{i,c}}, \label{eq:loss_CE}
\end{align}
where $\mbf{x}_{i}$ is the feature vector, $C$ the set of all classes, $c_i$ the true class of instance $\mbf{x}_{i}$, $\theta_{c}$ the linear classifier associated with class $c$, $n_c$ the number of instances of class $c$ in the training set and $\eta_{i,c}$ the logit associated with class $c$. This formulation greatly improves performance on a balanced test set. Unless otherwise noted, cross-entropy losses in the following are corrected with Balanced Softmax.

\subsection{Supervised Contrastive Learning}
Supervised Contrastive Learning (SCL) \cite{khosla2020supervised} is a modification of standard contrastive loss that considers all samples of the same class in a batch to the positives. The supervised contrastive loss for instance $i$ has the form
\begin{align}
    \mathcal{L}_{i}^{\mrm{SCL}} &= -\frac{1}{\abs{P_{i}^{-}}} \sum_{j \in P_{i}^{-}} \log \frac{\exp \left(\mbf{z}_{i} \cdot \mbf{z}_{j} / \tau \right)}{\sum\limits_{k \in A^{-}} \exp \left(\mbf{z}_{i} \cdot \mbf{z}_{k} / \tau \right)}, \label{eq:loss_SCL}
\end{align}
where $\mbf{z}_{i}$ is the contrastive embedding vector for the $i^{th}$ instance, $P_{i}^{-}$ is the set of indices of the instances with the same class as instance $i$ excluding instance $i$, $\abs{P_{i}^{-}}$ is its cardinality, $A^{-}$ the set of all indices in the batch and queue excluding $i$, and $\tau$ is a temperature scaling parameter. Because contrastive losses tend to learn more reliably with additional negative examples, we keep a memory queue of past image features generated from a second copy of the encoder following the MoCo framework \cite{he2020momentum}. Both $P_i^{-}$ and $A^{-}$ contain examples from the current batch and the momentum queue. 

The contrastive embedding vectors $\mbf{z}_{\ell}$ are given as
\begin{align*}
    \mbf{z}_{\ell} &= \frac{h\left(\mbf{x}_{\ell}\right)}{\norm{h\left(\mbf{x}_{\ell}\right)}_2},
\end{align*}
where $h$ is a MLP projection layer as per SimCLR \cite{chen2020simple}.

\subsection{Combined Losses}
As discussed in Section~\ref{sec:rel_cl}, methods like Hybrid-SC \cite{wang2021contrastive} and PaCo \cite{cui2021parametric} combine both a cross-entropy and a supervised contrastive loss. 

\subsubsection{Summed Loss} \label{sec:sum_loss} We first consider a scalar-weighted sum of both losses given by eqs. \eqref{eq:loss_CE} and \eqref{eq:loss_SCL}, similar to Hybrid-SC \cite{wang2021contrastive} but with fixed weights
\begin{align}
    % \mathcal{L}^{\mrm{sum}}_i &= -\lambda^{\mrm{CE}} \log \frac{n_{c_i} \exp \left(\mbf{x}_{i}^{\trans}\mbs{\theta}_{c_i} \right)}{\sum_{c \in C} n_{c} \exp \left(\mbf{x}_{i}^{\trans}\mbs{\theta}_{c} \right)} \\ \nonumber
    %  & \quad -\frac{\lambda^{\mrm{SCL}}}{\abs{P_{i}}} \sum_{j \in P_{i} \backslash i} \log \frac{\exp \left(\mbf{z}_{i} \cdot \mbf{z}_{j} / \tau \right)}{\sum_{k \in A \backslash i} \exp \left(\mbf{z}_{i} \cdot \mbf{z}_{k} / \tau \right)}
    \mathcal{L}^{\mrm{sum}}_i &= \lambda^{\mrm{CE}} \mathcal{L}^{\mrm{CE}}_i + \lambda^{\mrm{SCL}} \mathcal{L}^{\mrm{SCL}}_i. \label{eq:loss_sum} 
\end{align}
In practice, this simple summed formulation has a limited effect on performance, with final results similar to using only the cross-entropy term. We discuss this further in Section~\ref{sec:abl_loss_choice}.

\subsubsection{PaCo Loss} Another approach, presented in PaCo \cite{cui2021parametric}, is to treat the cross-entropy logits as additional positive and negative samples in the contrastive loss. The loss in PaCo has the form
\begin{align}
    \mathcal{L}^{\mrm{PaCo}}_i &=- \gamma_i \Bigg( \alpha \sum_{j \in P_i^-} \log \frac{\exp \left(\mbf{z}_{i} \cdot \mbf{z}_{j} / \tau \right)}{\sum\limits_{k \in A^-} \exp \left(\mbf{z}_{i} \cdot \mbf{z}_{k} / \tau \right)+ \sum\limits_{c \in C}  \eta_{i,c}} \nonumber \\ 
    & \quad + \beta \log \frac{\eta_{i,c_i}}{\sum\limits_{k \in A^-} \exp \left(\mbf{z}_{i} \cdot \mbf{z}_{k} / \tau \right) + \sum\limits_{c \in C}  \eta_{i,c}} \Bigg) \label{eq:loss_PaCo}, \\
    \gamma_i &=  \frac{1}{\alpha \abs{P_{i}^{-}} +\beta} \nonumber,
\end{align}
where $\alpha$ and $\beta$ control the relative importance of the contrastive and cross-entropy positives. The cross-entropy and contrastive terms have the same definitions as in eq. \eqref{eq:loss_CE} and eq. \eqref{eq:loss_SCL} respectively. The numerator term is a weighted sum between a cross-entropy and a contrastive term from the other positive instances. This is then normalized by the sum of all positive and negative terms in both losses. Because each sample has only a single positive logit in the cross-entropy but can have many positives in the contrastive loss, the summed loss skews towards the contrastive term as the number of positive examples in the batch and momentum queue $\abs{P_i^-}$ increases. 
% This means that the network optimizes for the contrastive objective for classes with many samples and for the cross-entropy term for classes with few instances.

\subsubsection{CIBL Loss} \label{sec:cibl_loss} Inspired by the PaCO loss above, we consider a new class instance balanced loss (CIBL) that explicitly considers the sum of a cross-entropy and a contrastive term, with relative importance proportional to the number of instances in the batch and queue
\begin{align}
    \mathcal{L}_i^{\mrm{CIBL}} &=-\frac{1}{\lambda^{\mrm{CE}} + \lambda^{\mrm{SCL}} \abs{P_i^-}} \Bigg( \lambda^{\mrm{CE}} \log \frac{\eta_{i,c_i}}{\sum\limits_{c \in C} \eta_{i,c}} \nonumber \\ 
    & \quad + \lambda^{\mrm{SCL}} \sum_{j \in P_{i}^-} \log \frac{\exp \left(\mbf{z}_{i} \cdot \mbf{z}_{j} / \tau \right)}{\sum\limits_{k \in A^-} \exp \left(\mbf{z}_{i} \cdot \mbf{z}_{k} / \tau \right)}\Bigg). \label{eq:loss_CIBL}
\end{align}
Figure~\ref{fig:schematic_CIBL} presents a schematic of the CIBL formulation.

In contrast to the PaCo formulation in eq. \eqref{eq:loss_PaCo}, the cross-entropy and contrastive terms are uncoupled in CIBL and only dependent on the number of other positive samples in the batch and queue $\abs{P_i^-}$. Because of this, it remains a weighted sum of two independent losses. This also means that we can change the formulation of either loss with fewer repercussions on the other. CIBL is compared to PaCo and other methods in Section~\ref{sec:results}. A comparison between the baseline given in eq. \eqref{eq:loss_CE}, the simple summed loss given by eq. \eqref{eq:loss_sum} and the proposed CIBL loss given by eq. \eqref{eq:loss_CIBL} is given in Section~\ref{sec:abl_loss_choice}.

\subsection{Logit Adjustment for Cosine Classifiers}
Recent state-of-the-art methods \cite{zhu2022balanced, xu2022constructing} use a normalized classifier for the cross-entropy and obtain improved performance. Here, we also implement the cross-entropy loss with a cosine distance to see if this improvement is also observed, giving us the normalized CIBL (NCIBL). This gives a normalized cross-entropy (NCE) loss of the form
\begin{align}
    \mathcal{L}_i^{\mrm{NCE}} &= -\log \frac{n_{c_i} \exp \left(\mrm{sim}\left(\mbf{x}_{i},\mbs{\theta}_{c_i}\right) /\gamma \right)}{\sum_{c \in C} n_{c} \exp \left( \mrm{sim}\left(\mbf{x}_{i},\mbs{\theta}_{c} \right) /\gamma \right)}  \label{eq:loss_norm}, \\
    \mrm{sim}\left(\mbf{x}_{i},\mbs{\theta}_{c_i}\right) &= \frac{\mbf{x}_{i}^{\trans} \mbs{\theta}_{c_i}}{\norm{\mbf{x}_{i}}_2 \norm{\mbs{\theta}_{c_i}}_2}, \nonumber
\end{align}
where $n_{c}$, $\mbf{x}_{i}$, $\theta_{c}$ are defined as in eq \eqref{eq:loss_CE} and where $\gamma$ is a temperature parameter that scales the sharpness of the softmax and controls how sensitive the loss is towards the largest terms. Note that because the magnitude of the temperature directly affects the magnitude of the logits, the relative effects of the added logit adjustment depend on temperature. This is obvious if the logit adjustment term $n_c$ is moved inside the exponential, acting as a margin on both the positive and negative logits.
\begin{align}
    \mathcal{L}_i^{\mrm{NCE}} &= -\log \frac{ \exp \left(\mrm{sim}\left(\mbf{x}_{i},\mbs{\theta}_{c_i}\right) /\gamma + \log n_{c_i} \right)}{\sum_{c \in C} \exp \left( \mrm{sim}\left(\mbf{x}_{i},\mbs{\theta}_{c} \right) /\gamma + \log n_{c} \right)}  \label{eq:loss_margin},
\end{align}
In practice, the performance of NCIBL is sensitive to the temperature value $\gamma$ used, but a reasonable range of values gives results that are better than the baseline of pure cross-entropy. Section~\ref{sec:abl_temp} presents a discussion of this. 

In CIBL, pulling the number of class instances $n_{c}$ into the exponential function as in eq. \eqref{eq:loss_margin} adds a margin $\log \left(n_{c}\right)$ that has a constant magnitude for all instances of a given class, regardless of the original logit $\mbf{x}_{i}^{\trans}\mbs{\theta}_{c}$. In NCIBL however, the effect of the added margin depends on the temperature hyperparameter $\gamma$. Moreover, it cannot be interpreted as a fixed margin on the angle between $\mbf{x}_{i}$ and $\mbs{\theta}_{c}$. In the related task of face recognition, ArcFace \cite{deng2019arcface} showed that a constant angular margin, with the margin adjustment done inside the similarity $\mrm{sim}$ function, was beneficial over having the margin added on the output of the similarity as done here, but further analysis of this result in the context of long-tailed image classification is outside the scope of this work.

%------------------------------------------------------------------------
\section{Experiments}
We compare the performance of CIBL and NCIBL against other current state-of-the-art methods on two standard long-tailed datasets: CIFAR-100-LT \cite{cao2019cifarlLT} and ImageNet-LT \cite{liu2019largeIN-LT-PL-LT}. We follow the standard evaluation protocol by testing on a class-balanced test set. The total classification accuracy is usually considered the primary metric of performance, but we also report the mean class-wise accuracy for three subsets of classes when testing on ImageNet: The $Many$ category includes classes with $n > 100$ images, $Medium$ those with $100 \geq n > 20$ and $Few$ those with $n \leq 20$. While we rank the different methods according to the average accuracy, we are also interested in identifying if the proposed approach improves performance on the tail classes specifically.

\subsection{Datasets}
\subsubsection{Long-Tailed CIFAR-100} CIFAR-100-LT is a subsampled version of the original CIFAR-100 dataset \cite{krizhevsky2009cifar}, which is composed of 50,000 training and 10,000 validation images of size $32 \times 32$, equally divided into 100 classes. The long-tailed version of the training dataset is subsampled with an exponential factor, and we present results for imbalance factors of 100, 50 and 10.

\subsubsection{Long-Tailed ImageNet} ImageNet-LT is a subsampling of the original ImageNet 2012 \cite{ILSVRC15} following a Pareto distribution, with 1000 classes having from 1280 to 5 instances in the training set. The test set is balanced, with 50 images per class. The original images have varying dimensions, but all of them are cropped for testing and training to a size of $224 \times 224$, with the crop being centred for the test images.

\subsection{Implementation Details} \label{sec:implementation}
To allow a fair comparison with PaCo \cite{cui2021parametric}, which is the closest method to our proposed CIBL, we generally follow the same training setups and schedules. This means that the supervised head uses logit adjustment following Balanced Softmax \cite{ren2020balanced} as described in Section~\ref{sec:log_adj} above and that the supervised contrastive loss is based on the MoCov2 \cite{chen2020improved} approach. All the training is done end-to-end on a single Nvidia 3090 GPU using SGD with a momentum of 0.9.

\subsubsection{CIFAR-100-LT} \label{sec:impl_cifar} We use ResNet-32 as the backbone. Note that this is the smaller version of ResNet as per the test setting in LDAM \cite{cao2019cifarlLT}. Following PaCo, we augment the main branch with Cutout \cite{devries2017improved} and AutoAugment \cite{cubuk2018autoaugment} and the second momentum branch with SimAugment \cite{chen2020simple}. The class instance balance parameters are set to $\lambda^{\mrm{CE}} = 1.0$ and $\lambda^{\mrm{SCL}} = 0.03$ respectively. We train for 400 epochs, with a learning rate of $0.1$, decaying to 0.01 at 320 and 0.001 at 360 epochs, and with a linear warmup for the first 10 epochs. We use a batch size of 128, a MoCo queue length of 1024 and a contrastive temperature of $\tau=0.05$. For NCIBL, we use a cross-entropy temperature of $\gamma = 0.05$.

\subsubsection{ImageNet-LT} \label{sec:impl_imagenet} We use ResNet-50 as the backbone. We augment the main branch with RandAugment \cite{cubuk2020randaugment} and the momentum branch with SimAugment. The class instance balance parameters are set to $\lambda^{\mrm{CE}} = 1.0$ and $\lambda^{\mrm{SCL}} = 0.05$ respectively. We present results for a training time of 400 epochs for CIBL, following PaCo. However, this schedule overfits when using NCIBL, and so we present results for 200 epochs instead for NCIBL. See Section~\ref{sec:overfit} for further discussion. For CIBL, we use a learning rate of 0.02, and for NCIBL, a learning rate of 0.04. Both decay to 0 using a cosine schedule and a linear warmup period of 10 epochs. We use a batch size of 128, a MoCo queue of 8192 and a contrastive temperature of $\tau=0.2$, again following the implementation in PaCo. NCIBL uses a temperature parameter of $\gamma=0.05$.

\subsection{Results} \label{sec:results}
\subsubsection{CIFAR-100-LT} Table~\ref{tab:res_cifar} presents the results on CIFAR-100 for our proposed CIBL method with and without the normalized classifier. Here, both proposed approaches outperform the baselines Balanced Softmax and PaCo and all the other methods for all the considered imbalance ratios. NCIBL performs similarly for an imbalance ratio of 100 but underperforms for smaller imbalances ratios.

\subsubsection{ImageNet-LT} Table~\ref{tab:res_imagenet} presents the results on ImageNet-LT for CIBL and NCIBL. Here, our approaches perform similarly to PaCo, and both proposed methods outperform all other methods. As on CIFAR-100, NCIBL has slightly lower accuracy than CIBL on Imagenet-LT, but the differences of $-0.1\%$ are likely not significant. However, NCIBL has $-1.5\%$ lower accuracy on the tail classes compared to CIBL and so presents a significantly less balanced performance. This is due to overfitting on the tail data, which is discussed in additional detail in Section~\ref{sec:overfit} below.

\begin{table}[t]
\centering
\begin{tabular}{c|ccc}
\hline \rule{0pt}{8pt}\text { Methods } & 100 & 50 & 10  \\
\hline \rule{0pt}{8pt} 
Focal \cite{lin2017focal}          & 38.4 & 44.3 & 55.8 \\
LDAM-DRW \cite{cao2019cifarlLT}      & 42.0   & 46.6 & 58.7 \\
TSC \cite{li2022targeted}            & 43.8 & 47.4 & 59.0  \\
Hybrid-SC \cite{wang2021contrastive}     & 46.7 & 51.9 & 63.1 \\
MisLAS  \cite{zhong2021improving}       & 47.0     &  52.3    &  63.2    \\
MetaSaug-LDAM \cite{li2021metasaug}  & 48.0   & 52.3 & 61.3  \\ 
ResLT \cite{cui2022reslt}         & 48.2 & 52.7 & 62.0   \\
RIDE (4 experts) \cite{wang2020long} & 49.1 &   -   &  -    \\
Balanced Softmax \cite{ren2020balanced}        & 50.8 &   -   & 63.0   \\
PaCo \cite{cui2021parametric}          & 52.0   &   -   &   -   \\ \hline
\rule{0pt}{8pt} CIBL           & \textbf{52.6} & \textbf{56.8} & \textbf{65.8} \\
NCIBL          & \underline{52.4} & \underline{56.4} & \underline{65.3} \\
\hline
\end{tabular}
\caption{Accuracy on CIFAR-100-LT. We present results for imbalance ratios of 100, 50 and 10. Bold values are the best, underlined second best.}
\label{tab:res_cifar}
\end{table}

\begin{table}[t]
\centering
\begin{tabular}{c|ccc|c}
\hline \rule{0pt}{8pt}  Methods & Many & Medium &  Few & All \\
\hline \rule{0pt}{8pt}
$\tau$-norm \cite{kang2019decoupling}     & 56.6 & 44.2 & 27.4 & 46.7 \\
KCL \cite{kang2021exploring}             & 61.8 & 49.4 & 30.9 & 51.5 \\
TSC \cite{li2022targeted}             & 63.5 & 49.7 & 30.4 & 52.4 \\
MisLAS \cite{zhong2021improving}           &  -    &   -   &   -   &   52.7   \\
DisAlign \cite{zhang2021distribution}       & 61.3 & 52.2 & 31.4 & 52.9 \\
Balanced Softmax$^{\dagger}$ \cite{ren2020balanced}     &   -   &    -  &  -    & 55.0  \\
vMF \cite{wang2022towards}            &   65.1   &   52.8   &   34.2  & 55.0   \\
RIDE (4 experts) \cite{wang2020long} &  -    &   -   &    -  & 55.4 \\
BCL \cite{zhu2022balanced}            &   -   &   -   &   -   & 56.0   \\
DLSA \cite{xu2022constructing}            &   64.6   &   54.9   &   \textbf{41.8}   & 56.9   \\
PaCo \cite{cui2021parametric}            &   -   &   -   &   -   & \textbf{57.0}   \\
\hline
CIBL             & \underline{65.4} & \textbf{55.2} & \underline{39.4} & \textbf{57.0}   \\
NCIBL            & \textbf{65.8} & \textbf{55.2} & 37.9 & 56.9 \\
\hline    
\end{tabular}
\caption{Accuracy on ImageNet-LT. $\dagger$ denotes results provided by \cite{cui2021parametric}. Bold values are the best, underlined second best.}
\label{tab:res_imagenet}
\end{table}

\subsection{A Discussion on Training Length}\label{sec:overfit}
In image classification, increasing the number of training epochs generally leads to improved performance. However, there are diminishing returns on extended training and the eventual risk of overfitting to the training data. However, the number of training epochs required to reach overfitting will depend on the method used.

Table~\ref{tab:n_epochs} compares the performances on ImageNet-LT of CIBL and NCIBL as the number of training epochs is increased. The implementation is the same as in Section~\ref{sec:impl_imagenet}, except NCIBL at 100 epochs is trained with a learning rate of 0.02. For both CIFAR-100 and ImageNet, increasing the learning rate when close to the best operating point for maximum accuracy tends to decrease overfitting with respect to the test set. Thus, we increase the learning rate for the longer training schedules of 200 and 400 epochs for NCIBL.

Here, we see that NCIBL has higher accuracy for shorter training times, providing an improvement of $+3.3\%$ at 100 epochs. At 200 epochs, the gain is still $+1.4\%$, and only $-0.1\%$ behind CIBL at 400 epochs. However, note that the performance on rare classes has dropped compared to NCIBL at 100 epochs, indicating the method overfitting these classes. For the 400 epoch run, NCIBL performs worse than CIBL by $-0.6\%$ and worse than the NCIBL run at 200 epochs by $-0.5\%$, and there is a loss in performance on both the medium and tail classes, again indicating overfit. Further increases in epochs are likely to continue to reduce performance. In the end, the normalized formulation allows a significantly faster training, with similar performance to CIBL with only half the number of epochs, but not necessarily a better one.

\begin{figure*}[!ht]
    \centering
	\includegraphics[width=1.00\textwidth]{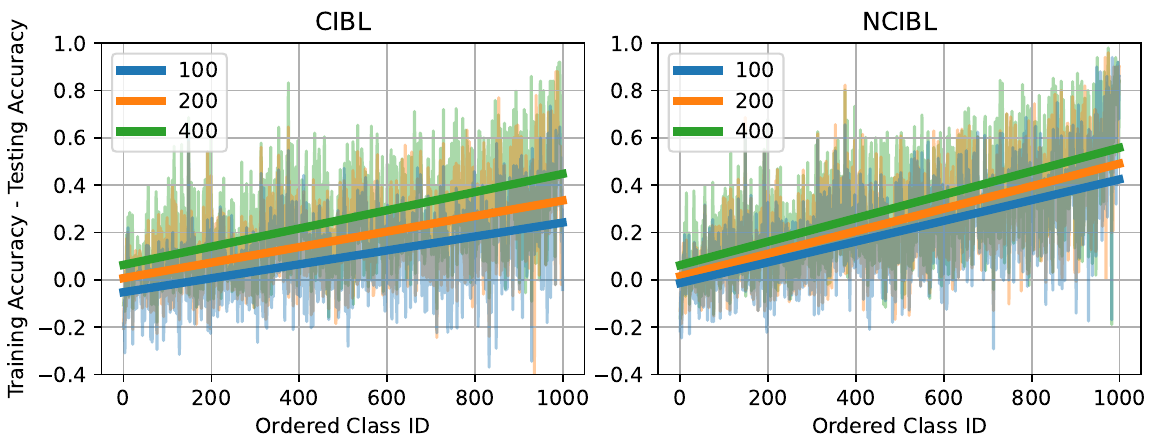}
	\caption{Class accuracy overfitting on ImageNet-LT for different training lengths. The classes are ordered in decreasing order of frequency in the training dataset. The bold lines are linear fits of the data.}
	\label{fig:overfit}
\end{figure*}

Finally, we see that as the number of training epochs increases, improvements in accuracy on the training data are less transferable to the test set. Figure~\ref{fig:overfit} presents the difference between training and testing accuracies at the last training epoch for all classes, ordered by test set frequency, and linear fits to the data. For both CIBL and NCIBL, the increase in training epochs leads to a higher y-intercept and a steeper slope on the linear fits. Overfitting tends to worsen as the number of epochs increases, and this result is accentuated on tail classes.

\begin{table}[t]
\centering
\begin{tabular}{cc|ccc|c|c}
\hline \rule{0pt}{8pt}  Loss  & Epochs & Many  &  Medium  &  Few  &  All   & Train  \\ 
\hline \rule{0pt}{8pt}CIBL & 100 & 58.5 & 51.3 & 31.7 & 51.4 & 57.9  \\ 
  CIBL & 200 & 62.8 & 54.7 & 37.7 & 55.5 & 69.1  \\ 
  CIBL & 400 & 65.4 & 55.2 & 39.4 & 57.0 & 78.8  \\ 
  \hline \rule{0pt}{8pt}NCIBL & 100 & 61.8 & 53.5 & 38.9 & 54.7 & 69.4 \\ 
  NCIBL & 200 & 65.8 & 55.2 & \cellcolor{pink!80}37.9 & 56.9 & 77.2 \\ 
  NCIBL & 400 & 67.2 & \cellcolor{pink!80}53.4 & \cellcolor{pink!80}36.4 & \cellcolor{pink!80}56.4 & 83.4 \\
\hline
\end{tabular}
\caption{Effect of the number of training epochs on ImageNet-LT. Highlighted cells indicate worse performance for increased training epochs.}
\label{tab:n_epochs}
\end{table}

\subsection{Ablations on the Choice of Loss Formulation} \label{sec:abl_loss_choice}
We compare the baseline cross-entropy loss with and without logit adjustment given by eq. \eqref{eq:loss_CE}, the summed loss given by eq. \eqref{eq:loss_sum} and our CIBL loss given by eq. \eqref{eq:loss_CIBL} to identify the effects of the contrastive scaling term $\lambda^{\mrm{SCL}}$ and of our specific formulation. Results are presented in Table~\ref{tab:abl_scl_lambda} and are obtained using the same test setting as the experiments on CIFAR-100-LT discussed in Section~\ref{sec:impl_cifar}. 

\begin{table}[th]
\centering
\begin{tabular}{cc|ccc|c|c}
\hline \rule{0pt}{10pt}  Loss  & $\lambda^{\mrm{SCL}}$ &  Many  &  Medium  &  Few  &  All   & Train  \\ 
\hline  \rule{0pt}{8pt}CE$^{\ddagger}$ & - & \textbf{75.3} & 46.9 & 9.8 & 45.8 & 78.4  \\
  CE & - & 67.6 & 52.1 & 30.5 & 51.0 & 72.1  \\ 
  \hline  \rule{0pt}{8pt}Sum & 0.01 & \underline{69.7} & 53.2 & 29.0 & 51.6 & 72.9  \\ 
  Sum  & 0.1 & 69.3 & 53.5 & 30.0 & 52.0 & 73.5  \\ 
  Sum  & 0.5 & 68.3 & 53.3 & 31.3 & 51.9 & 72.4  \\ 
  Sum  & 1.0 & 68.2 & 53.7 & 31.5 & 52.1 & 70.9  \\ 
  Sum  & 2.0 & 66.9 & 53.1 & 29.8 & 51.0 & 68.7  \\ 
  \hline  \rule{0pt}{8pt}CIBL  & 0.01 & 67.2 & 53.5 & 33.2 & \underline{52.4} & 70.5  \\ 
  CIBL & 0.03 & 65.1 & 53.2 & 36.4 & \textbf{52.6} & 65.2  \\ 
  CIBL & 0.05 & 61.5 & \underline{53.9} & \underline{36.8} & 51.4 & 51.1  \\
  CIBL & 0.10 & 57.7 & \textbf{54.3} & \textbf{38.4} & 50.7 & 50.4  \\
\hline
\end{tabular}
\caption{Comparison between different combined losses on CIFAR-100-LT. We use larger steps in $\lambda^{\mrm{SCL}}$ for the summed loss to observe significant changes with respect to the cross-entropy method. The CE$^\ddagger$ method is implemented without logit adjustment. Bold values are the best, underlined second best.}
\label{tab:abl_scl_lambda}
\end{table}

We see that including the contrastive loss can improve performance for both formulations as long as $\lambda^{\mrm{SCL}}$ is not too large, after which the average performance degrades. For CIBL, increasing $\lambda^{\mrm{SCL}}$ also transfers performance from frequent classes towards rare classes and reduces the overfit on the training set. These effects are also seen with the summed loss, but the magnitudes are much smaller. This is likely because in CIBL, the relative importance of the contrastive term is scaled by the class frequency in a batch $\abs{P_i^-}$. This means that changes in $\lambda^{\mrm{SCL}}$ mainly affect the loss on frequent classes, in contrast to the loss for rare classes, which is always dominated by cross-entropy. This difference in training objective allows the network to learn a more balanced representation, unlike simply using a sum of the losses with the same weights for all instances. Finally, the performance on the tail classes increases with the magnitude of $\lambda^{\mrm{SCL}}$, regardless of the change in average accuracy. This means that the end user can sacrifice some average performance for rare class performance if desired by increasing $\lambda^{\mrm{SCL}}$.

\subsection{Ablations on the Temperature Scaling $\gamma$ for NCIBL} \label{sec:abl_temp}
Table~\ref{tab:abl_tau_cos} presents the performance of the NCIBL loss on CIFAR-100 for different values of $\gamma$. It has been shown \cite{kornblith2021better} that decreasing the temperature tends to improve the fit on the dataset at the cost of a loss of generalizability. Thus, we expect the average performance to increase as the temperature decreases, but the performance on rare classes may also decrease as overfitting worsens. Here, the network trains poorly for large values of $\gamma$ (1, 0.2), with training and testing accuracies significantly under the unnormalized cross-entropy baseline. As the value of $\gamma$ gets smaller, the testing accuracy increases to a maximum at $\gamma = 0.05$. At $\gamma = 0.025$, the performance is still better than the baseline, but the test accuracy is worse than at $\gamma = 0.05$, particularly for the rare classes where it drops by $-3.5\%$. On the other hand, the training accuracy continues to increase, indicating overfitting. The optimal temperature for average accuracy is also where the performance on the rarest classes is maximized. Note also that the optimal temperature $\gamma$ of 0.05 is the same as the contrastive temperature $\tau$ used in the contrastive loss for CIFAR as described in Section~\ref{sec:implementation}. While the performance of NCIBL is dependent on $\gamma$, this parameter takes on values similar to other temperature scaling parameters used in normalized losses.

\begin{table}[th]
\centering
\begin{tabular}{cc|ccc|c|c}
\hline \rule{0pt}{8pt}  Loss  & $\gamma$ & Many  &  Medium  &  Few  &  All   & Train  \\ 
\hline \rule{0pt}{8pt}CE & - & \textbf{67.6} & 52.1 & 30.5 & 51.0 & 72.1  \\ 
  NCIBL & 1 & 27.5 & 30.9 & 17.5 & 25.7  & 23.2  \\ 
  NCIBL & 0.2 & 63.9 & 48.1 & 18.7 & 44.8 & 59.0  \\ 
  NCIBL & 0.1 & 63.4 & 53.0 & \underline{32.5} & 50.6 & 65.4 \\ 
  NCIBL & 0.05 & 64.7 & \textbf{55.6} & \textbf{34.3} & \textbf{52.4} & 68.6 \\ 
  NCIBL & 0.025 & \underline{66.4} & \underline{54.7} & 30.8 & \underline{51.9} & 70.6 \\
\hline
\end{tabular}
\caption{Effects temperature magnitude $\gamma$ for NCIBL on CIFAR-100-LT. Bold values are the best, underlined second best.}
\label{tab:abl_tau_cos}
\end{table}

\section{Conclusion}
In this paper, we propose Class Instance Balanced Loss (CIBL), a method of combining cross-entropy and contrastive losses to improve long-tailed image classification. Our analysis shows that by forcing the loss on frequent class samples to favour the contrastive loss over the cross-entropy loss, we obtain a more accurate and more balanced final classifier compared to using a fixed weighting for all classes while also enabling a principled approach to the tradeoff between head and tail performance, which can be tuned using a single parameter. We also show that replacing the linear classifier with a cosine classifier allows us to train a network with similar average performance in significantly fewer epochs. Our method obtains competitive results compared to state-of-the-art methods on CIFAR-100-LT and ImageNet-LT.

\bibliographystyle{IEEEtran}
% \bibliographystyle{unsrt}
% This one
\bibliography{egbib.bib}     

\begin{thebibliography}{10}
\providecommand{\url}[1]{#1}
\csname url@samestyle\endcsname
\providecommand{\newblock}{\relax}
\providecommand{\bibinfo}[2]{#2}
\providecommand{\BIBentrySTDinterwordspacing}{\spaceskip=0pt\relax}
\providecommand{\BIBentryALTinterwordstretchfactor}{4}
\providecommand{\BIBentryALTinterwordspacing}{\spaceskip=\fontdimen2\font plus
\BIBentryALTinterwordstretchfactor\fontdimen3\font minus
  \fontdimen4\font\relax}
\providecommand{\BIBforeignlanguage}[2]{{%
\expandafter\ifx\csname l@#1\endcsname\relax
\typeout{** WARNING: IEEEtran.bst: No hyphenation pattern has been}%
\typeout{** loaded for the language `#1'. Using the pattern for}%
\typeout{** the default language instead.}%
\else
\language=\csname l@#1\endcsname
\fi
#2}}
\providecommand{\BIBdecl}{\relax}
\BIBdecl

\bibitem{caesar2020nuscenes}
H.~Caesar, V.~Bankiti, A.~H. Lang, S.~Vora, V.~E. Liong, Q.~Xu, A.~Krishnan,
  Y.~Pan, G.~Baldan, and O.~Beijbom, ``Nuscenes: A multimodal dataset for
  autonomous driving,'' in \emph{Proc. IEEE/CVF Conference on Computer Vision
  and Pattern Recognition (CVPR)}, 2020.

\bibitem{he2009learning}
H.~He and E.~A. Garcia, ``Learning from imbalanced data,'' \emph{IEEE
  Transactions on Knowledge and Data Engineering}, vol.~21, no.~9, pp.
  1263--1284, 2009.

\bibitem{huang2016learning}
C.~Huang, Y.~Li, C.~C. Loy, and X.~Tang, ``Learning deep representation for
  imbalanced classification,'' in \emph{Proc. IEEE/CVF Conference on Computer
  Vision and Pattern Recognition (CVPR)}, 2016.

\bibitem{mahajan2018exploring}
D.~Mahajan, R.~Girshick, V.~Ramanathan, K.~He, M.~Paluri, Y.~Li, A.~Bharambe,
  and L.~Van Der~Maaten, ``Exploring the limits of weakly supervised
  pretraining,'' in \emph{Proc. European Conference on Computer Vision (ECCV)},
  2018.

\bibitem{khan2017cost}
S.~H. Khan, M.~Hayat, M.~Bennamoun, F.~A. Sohel, and R.~Togneri,
  ``Cost-sensitive learning of deep feature representations from imbalanced
  data,'' \emph{IEEE Transactions on Neural Networks and Learning Systems},
  vol.~29, no.~8, pp. 3573--3587, 2017.

\bibitem{cui2019class}
Y.~Cui, M.~Jia, T.-Y. Lin, Y.~Song, and S.~Belongie, ``Class-balanced loss
  based on effective number of samples,'' in \emph{Proc. IEEE/CVF Conference on
  Computer Vision and Pattern Recognition (CVPR)}, 2019.

\bibitem{buda2018systematic}
M.~Buda, A.~Maki, and M.~A. Mazurowski, ``A systematic study of the class
  imbalance problem in convolutional neural networks,'' \emph{Neural Networks},
  vol. 106, pp. 249--259, 2018.

\bibitem{yun2019cutmix}
S.~Yun, D.~Han, S.~J. Oh, S.~Chun, J.~Choe, and Y.~Yoo, ``Cutmix:
  Regularization strategy to train strong classifiers with localizable
  features,'' in \emph{Proc. IEEE/CVF International Conference on Computer
  Vision (ICCV)}, 2019.

\bibitem{li2020adversarial}
K.~Li, Y.~Zhang, K.~Li, and Y.~Fu, ``Adversarial feature hallucination networks
  for few-shot learning,'' in \emph{Proc. IEEE/CVF Conference on Computer
  Vision and Pattern Recognition (CVPR)}, 2020.

\bibitem{li2021metasaug}
S.~Li, K.~Gong, C.~H. Liu, Y.~Wang, F.~Qiao, and X.~Cheng, ``Metasaug: Meta
  semantic augmentation for long-tailed visual recognition,'' in \emph{Proc.
  IEEE/CVF Conference on Computer Vision and Pattern Recognition (CVPR)}, 2021.

\bibitem{cao2019cifarlLT}
K.~Cao, C.~Wei, A.~Gaidon, N.~Arechiga, and T.~Ma, ``Learning imbalanced
  datasets with label-distribution-aware margin loss,'' \emph{arXiv preprint
  arXiv:1906.07413}, 2019.

\bibitem{ren2020balanced}
J.~Ren, C.~Yu, X.~Ma, H.~Zhao, S.~Yi \emph{et~al.}, ``Balanced meta-softmax for
  long-tailed visual recognition,'' \emph{Advances in Neural Information
  Processing Systems}, vol.~33, pp. 4175--4186, 2020.

\bibitem{chen2020simple}
T.~Chen, S.~Kornblith, M.~Norouzi, and G.~Hinton, ``A simple framework for
  contrastive learning of visual representations,'' in \emph{International
  conference on machine learning}, 2020.

\bibitem{chen2020improved}
X.~Chen, H.~Fan, R.~Girshick, and K.~He, ``Improved baselines with momentum
  contrastive learning,'' \emph{arXiv preprint arXiv:2003.04297}, 2020.

\bibitem{kang2021exploring}
B.~Kang, Y.~Li, S.~Xie, Z.~Yuan, and J.~Feng, ``Exploring balanced feature
  spaces for representation learning,'' in \emph{International Conference on
  Learning Representations}, 2021.

\bibitem{wang2021contrastive}
P.~Wang, K.~Han, X.-S. Wei, L.~Zhang, and L.~Wang, ``Contrastive learning based
  hybrid networks for long-tailed image classification,'' in \emph{Proc.
  IEEE/CVF Conference on Computer Vision and Pattern Recognition (CVPR)}, 2021.

\bibitem{cui2021parametric}
J.~Cui, Z.~Zhong, S.~Liu, B.~Yu, and J.~Jia, ``Parametric contrastive
  learning,'' in \emph{Proc. IEEE/CVF International Conference on Computer
  Vision (ICCV)}, 2021.

\bibitem{zhu2022balanced}
J.~Zhu, Z.~Wang, J.~Chen, Y.-P.~P. Chen, and Y.-G. Jiang, ``Balanced
  contrastive learning for long-tailed visual recognition,'' in \emph{Proc.
  IEEE/CVF Conference on Computer Vision and Pattern Recognition (CVPR)}, 2022.

\bibitem{lin2017focal}
T.-Y. Lin, P.~Goyal, R.~Girshick, K.~He, and P.~Doll{\'a}r, ``Focal loss for
  dense object detection,'' in \emph{Proc. IEEE International Conference on
  Computer Vision (ICCV)}, 2017.

\bibitem{kang2019decoupling}
B.~Kang, S.~Xie, M.~Rohrbach, Z.~Yan, A.~Gordo, J.~Feng, and Y.~Kalantidis,
  ``Decoupling representation and classifier for long-tailed recognition,''
  \emph{arXiv preprint arXiv:1910.09217}, 2019.

\bibitem{zhang2021distribution}
S.~Zhang, Z.~Li, S.~Yan, X.~He, and J.~Sun, ``Distribution alignment: A unified
  framework for long-tail visual recognition,'' in \emph{Proc. IEEE/CVF
  Conference on Computer Vision and Pattern Recognition (CVPR)}, 2021.

\bibitem{xu2022constructing}
Y.~Xu, Y.-L. Li, J.~Li, and C.~Lu, ``Constructing balance from imbalance for
  long-tailed image recognition,'' in \emph{Proc. IEEE/CVF Conference on
  Computer Vision and Pattern Recognition (CVPR)}, 2022.

\bibitem{deng2019arcface}
J.~Deng, J.~Guo, N.~Xue, and S.~Zafeiriou, ``Arcface: Additive angular margin
  loss for deep face recognition,'' in \emph{Proc. IEEE/CVF Conference on
  Computer Vision and Pattern Recognition (CVPR)}, 2019.

\bibitem{menon2020long}
A.~K. Menon, S.~Jayaplana, A.~S. Rawat, H.~Jain, A.~Veit, and S.~Kumar,
  ``Long-tail learning via logit adjustment,'' \emph{arXiv preprint
  arXiv:2007.07314}, 2020.

\bibitem{zhao2022adaptive}
Y.~Zhao, W.~Chen, X.~Tan, K.~Huang, and J.~Zhu, ``Adaptive logit adjustment
  loss for long-tailed visual recognition,'' in \emph{Proc. AAAI Conference on
  Artificial Intelligence}, 2022.

\bibitem{kim2020m2m}
J.~Kim, J.~Jeong, and J.~Shin, ``M2m: Imbalanced classification via
  major-to-minor translation,'' in \emph{Proc. IEEE/CVF Conference on Computer
  Vision and Pattern Recognition (CVPR)}, 2020.

\bibitem{park2022majority}
S.~Park, Y.~Hong, B.~Heo, S.~Yun, and J.~Y. Choi, ``The majority can help the
  minority: Context-rich minority oversampling for long-tailed
  classification,'' in \emph{Proc. IEEE/CVF Conference on Computer Vision and
  Pattern Recognition (CVPR)}, 2022.

\bibitem{mullick2019generative}
S.~S. Mullick, S.~Datta, and S.~Das, ``Generative adversarial minority
  oversampling,'' in \emph{Proc. IEEE/CVF International Conference on Computer
  Vision (ICCV)}, 2019.

\bibitem{zada2022pure}
S.~Zada, I.~Benou, and M.~Irani, ``Pure noise to the rescue of insufficient
  data: Improving imbalanced classification by training on random noise
  images,'' in \emph{Proc. International Conference on Machine Learning
  (ICML)}, 2022.

\bibitem{wang2020long}
X.~Wang, L.~Lian, Z.~Miao, Z.~Liu, and S.~X. Yu, ``Long-tailed recognition by
  routing diverse distribution-aware experts,'' \emph{arXiv preprint
  arXiv:2010.01809}, 2020.

\bibitem{xiang2020learning}
L.~Xiang, G.~Ding, and J.~Han, ``Learning from multiple experts: Self-paced
  knowledge distillation for long-tailed classification,'' in \emph{Proc.
  European Conference on Computer Vision (ECCV)}, 2020.

\bibitem{li2022nested}
J.~Li, Z.~Tan, J.~Wan, Z.~Lei, and G.~Guo, ``Nested collaborative learning for
  long-tailed visual recognition,'' in \emph{Proc. IEEE/CVF Conference on
  Computer Vision and Pattern Recognition (CVPR)}, 2022.

\bibitem{cui2022reslt}
J.~Cui, S.~Liu, Z.~Tian, Z.~Zhong, and J.~Jia, ``Reslt: Residual learning for
  long-tailed recognition,'' \emph{IEEE Transactions on Pattern Analysis and
  Machine Intelligence}, 2022.

\bibitem{he2020momentum}
K.~He, H.~Fan, Y.~Wu, S.~Xie, and R.~Girshick, ``Momentum contrast for
  unsupervised visual representation learning,'' in \emph{Proc. IEEE/CVF
  Conference on Computer Vision and Pattern Recognition (CVPR)}, 2020.

\bibitem{yang2020rethinking}
Y.~Yang and Z.~Xu, ``Rethinking the value of labels for improving
  class-imbalanced learning,'' \emph{Advances in Neural Information Processing
  Systems}, vol.~33, 2020.

\bibitem{li2021self}
T.~Li, L.~Wang, and G.~Wu, ``Self supervision to distillation for long-tailed
  visual recognition,'' in \emph{Proc. of the IEEE/CVF International Conference
  on Computer Vision}, 2021.

\bibitem{khosla2020supervised}
P.~Khosla, P.~Teterwak, C.~Wang, A.~Sarna, Y.~Tian, P.~Isola, A.~Maschinot,
  C.~Liu, and D.~Krishnan, ``Supervised contrastive learning,'' \emph{Advances
  in Neural Information Processing Systems}, vol.~33, pp. 18\,661--18\,673,
  2020.

\bibitem{li2022targeted}
T.~Li, P.~Cao, Y.~Yuan, L.~Fan, Y.~Yang, R.~S. Feris, P.~Indyk, and D.~Katabi,
  ``Targeted supervised contrastive learning for long-tailed recognition,'' in
  \emph{Proc. IEEE/CVF Conference on Computer Vision and Pattern Recognition
  (CVPR)}, 2022.

\bibitem{liu2019largeIN-LT-PL-LT}
Z.~Liu, Z.~Miao, X.~Zhan, J.~Wang, B.~Gong, and S.~X. Yu, ``Large-scale
  long-tailed recognition in an open world,'' in \emph{Proc. IEEE/CVF
  Conference on Computer Vision and Pattern Recognition (CVPR)}, 2019.

\bibitem{krizhevsky2009cifar}
A.~Krizhevsky, G.~Hinton \emph{et~al.}, ``Learning multiple layers of features
  from tiny images,'' 2009.

\bibitem{ILSVRC15}
O.~Russakovsky, J.~Deng, H.~Su, J.~Krause, S.~Satheesh, S.~Ma, Z.~Huang,
  A.~Karpathy, A.~Khosla, M.~Bernstein, A.~C. Berg, and L.~Fei-Fei, ``{ImageNet
  Large Scale Visual Recognition Challenge},'' \emph{International Journal of
  Computer Vision (IJCV)}, vol. 115, no.~3, pp. 211--252, 2015.

\bibitem{devries2017improved}
T.~DeVries and G.~W. Taylor, ``Improved regularization of convolutional neural
  networks with cutout,'' \emph{arXiv preprint arXiv:1708.04552}, 2017.

\bibitem{cubuk2018autoaugment}
E.~D. Cubuk, B.~Zoph, D.~Mane, V.~Vasudevan, and Q.~V. Le, ``Autoaugment:
  Learning augmentation policies from data,'' \emph{arXiv preprint
  arXiv:1805.09501}, 2018.

\bibitem{cubuk2020randaugment}
E.~D. Cubuk, B.~Zoph, J.~Shlens, and Q.~V. Le, ``Randaugment: Practical
  automated data augmentation with a reduced search space,'' in \emph{Proc.
  IEEE/CVF Conference on Computer Vision and Pattern Recognition (CVPR)
  workshops}, 2020.

\bibitem{zhong2021improving}
Z.~Zhong, J.~Cui, S.~Liu, and J.~Jia, ``Improving calibration for long-tailed
  recognition,'' in \emph{Proc. IEEE/CVF Conference on Computer Vision and
  Pattern Recognition (CVPR)}, 2021.

\bibitem{wang2022towards}
H.~Wang, S.~Fu, X.~He, H.~Fang, Z.~Liu, and H.~Hu, ``Towards calibrated
  hyper-sphere representation via distribution overlap coefficient for
  long-tailed learning,'' in \emph{Proc. IEEE/CVF Conference on Computer Vision
  and Pattern Recognition (CVPR)}, 2022.

\bibitem{kornblith2021better}
S.~Kornblith, T.~Chen, H.~Lee, and M.~Norouzi, ``Why do better loss functions
  lead to less transferable features?'' \emph{Advances in Neural Information
  Processing Systems}, vol.~34, pp. 28\,648--28\,662, 2021.

\end{thebibliography}

\end{document}